\documentclass[10pt,twocolumn,letterpaper]{article}

\usepackage{iccv}
\usepackage{times}
\usepackage{epsfig}
\usepackage{graphicx}
\usepackage{amsmath}
\usepackage{amssymb}

\usepackage{multirow}
\usepackage{tabularx}
\usepackage{booktabs}
\usepackage{pifont}

\usepackage[dvipsnames]{xcolor}

\newcommand{\cmark}{\ding{51}}%
\newcommand{\xmark}{\ding{55}}%

\newcommand{\F}{\mathcal{F}}

\usepackage{algorithm} 
\usepackage{algorithmic}

\newcommand*\samethanks[1][\value{footnote}]{\footnotemark[#1]}

\usepackage{enumerate}
\definecolor{mygray}{gray}{0.7} 

\usepackage[caption=false]{subfig}

\usepackage[breaklinks=true,bookmarks=false]{hyperref}

\iccvfinalcopy 

\def\iccvPaperID{15} 

\ificcvfinal\fi

\begin{document}


\title{VLG-Net: Video-Language Graph Matching Network for Video Grounding}

{\author{
Mattia Soldan\thanks{equal contribution.}, 
\quad Mengmeng Xu \samethanks, 
\quad Sisi Qu \samethanks, 
\quad Jesper Tegner, 
\quad Bernard Ghanem\\
King Abdullah University of Science and Technology (KAUST), Thuwal, Saudi Arabia\\
{\tt\small \{mattia.soldan, mengmeng.xu, sisi.qu, jesper.tegner, bernard.ghanem\}@kaust.edu.sa}}}

\maketitle
\ificcvfinal\thispagestyle{empty}\fi

\begin{abstract}
Grounding language queries in videos aims at identifying the time interval (or moment) semantically relevant to a language query. The solution to this challenging task demands understanding videos' and queries' semantic content and the fine-grained reasoning about their multi-modal interactions. Our key idea is to recast this challenge into an algorithmic graph matching problem. Fueled by recent advances in Graph Neural Networks, we propose to leverage Graph Convolutional Networks to model video and textual information as well as their semantic alignment. To enable the mutual exchange of information across the modalities, we design a novel Video-Language Graph Matching Network (VLG-Net) to match video and query graphs. Core ingredients include representation graphs built atop video snippets and query tokens separately and used to model intra-modality relationships. A Graph Matching layer is adopted for cross-modal context modeling and multi-modal fusion. Finally, moment candidates are created using masked moment attention pooling by fusing the moment's enriched snippet features. We demonstrate superior performance over state-of-the-art grounding methods on three widely used datasets for temporal localization of moments in videos with language queries: ActivityNet-Captions, TACoS, and DiDeMo.
\end{abstract}
\vspace{-0.3cm}
\section{Introduction}\label{sec: intro}
Temporal action understanding is at the forefront of computer vision research. 
Hendricks~\etal~\cite{Hendricks_2017_ICCV} and Gao~\etal~\cite{Gao_2017_ICCV} recently introduced the task of temporally grounding language queries in videos as a generalization of the temporal action localization task, aiming to overcome the constraint of a predefined set of actions. 
This novel interdisciplinary task has gained momentum within the vision and language communities for its relevance and possible applications in video retrieval~\cite{dong2019dual, Shao_2018_ECCV, yu2018joint}, video question answering~\cite{huang2020location,lei2018tvqa}, human-computer interaction~\cite{zhu2020vision}, and video storytelling~\cite{gella2018dataset}. 
Enabling this fine-grained matching of language in videos can be adopted by professional video content creators during the editing process. For example, video editing often requires searching through many hours of raw, unlabelled video content for specific interesting highlights. Thus, the ability to retrieve such highlights through textual queries could provide a faster experience. 

\begin{figure}[t]
    \centering
    \includegraphics[trim={0cm 0cm 0cm 0cm},width=\linewidth,clip]{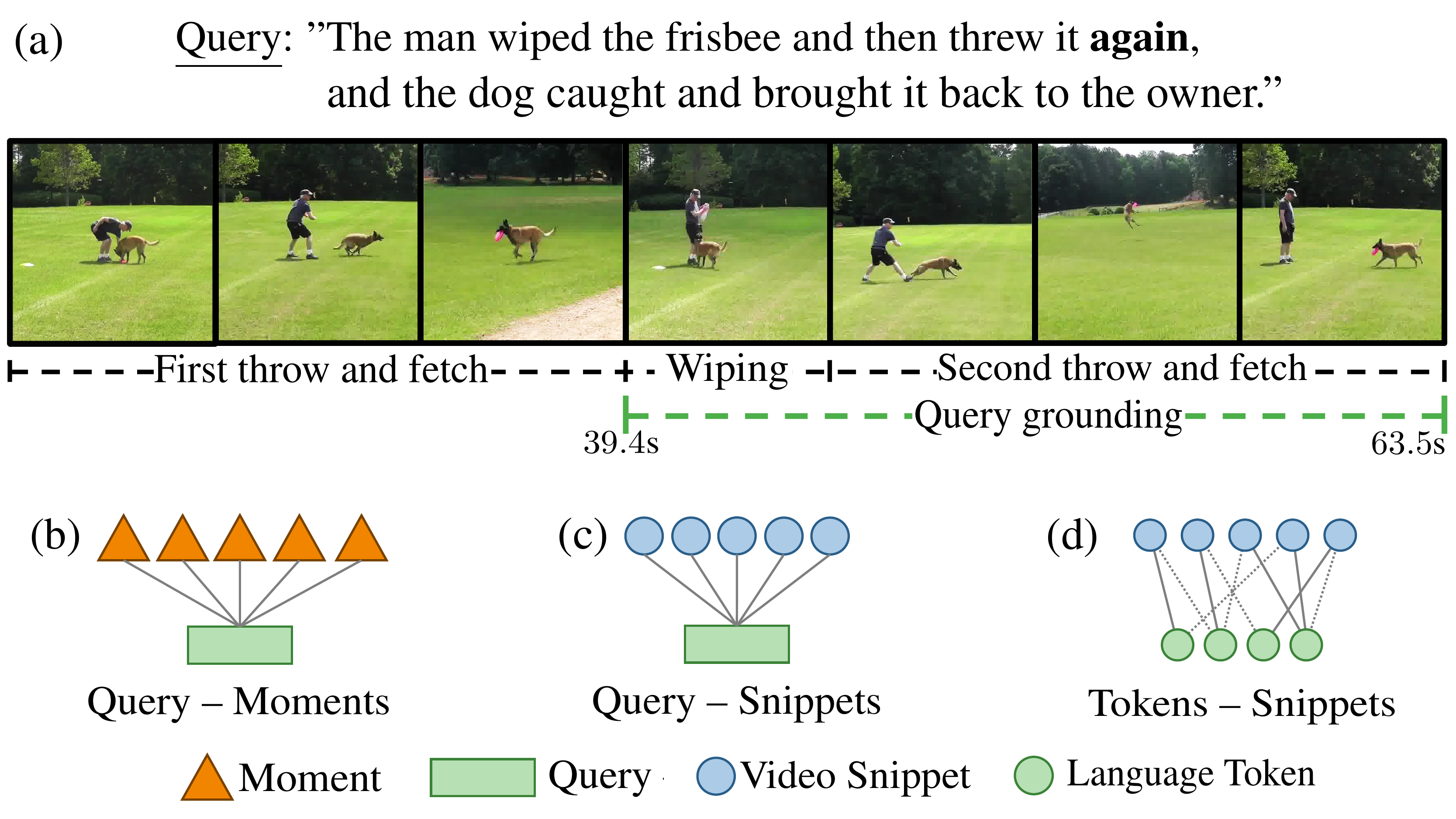}
    \caption{\textbf{Temporal video grounding task and multi-modality interaction schemes.} (a) A video grounding example showcasing the importance of fine-grained semantic understanding and  proper context modeling. 
    (b,c,d) Approaches for multi-modal interactions. We regard moments and queries as sequences of snippets and tokens respectively.
    We employ scheme (d), which allows for fine alignment by snippet-token matching. }
    \label{fig:intro}
    \vspace{-0.3cm}
\end{figure}

\indent Natural language grounding in videos inherits challenges from temporal action localization such as context modeling and candidate moment generation \cite{alwassel_2018_detad}. Semantic context is a fundamental cue necessary to boost the performance of localization methods~\cite{Dai_2017_ICCV, Gao_2017_ICCV_TURN,Lin_2018_ECCV, TSN_ECCV_2016, Xu_2020_CVPR}. To enrich video representation~\cite{Hendricks_2017_ICCV} adopted \textit{global-context}, which is moment independent, leading to sub-optimal performance. Conversely, a moment specific \textit{local-context}, defined as a moment's temporal neighbourhood, was used in~\cite{Gao_2017_ICCV, MAC_WACV_2019, jiang2019cross, 10.1145/3240508.3240549, 10.1007/978-3-030-00767-6_32}. 
In our view, \textit{non-local context} merits deeper analysis, since it has the potential to identify relevant information not restricted to the temporal neighbourhood within one data modality.
For example, in Fig.~\ref{fig:intro}(a), although ``\textit{First throw and fetch}'' is not in the temporal vicinity of ``\textit{Second throw and fetch}'', it is still semantically related with the target moment, showcasing the importance of non-local context modeling for video grounding. 

Moreover and as shown by the example, the free-form nature of the language modality introduces additional challenges. A model must understand the semantic content of both videos and language queries and reason about their multi-modal interactions.
Previous studies \cite{Gao_2017_ICCV, Ge_2019_WACV, 10.1007/978-3-030-00767-6_32} employ a cross-modal processing unit designed to jointly model text and visual features through simple operations such as element-wise addition, Hadamard product, and direct concatenation of a moment's representation and the query embedding.
A high-level overview of this multi-modal interaction scheme is presented in Fig.~\ref{fig:intro}(b).
Instead, recent works such as~\cite{Mun_2020_CVPR}, only employ the Hadamard product to fuse the multi-modal information at the query-snippet level. This scheme, depicted in Fig.~\ref{fig:intro}(c), can determine different correlations between a query and each video snippet, allowing for a finer fusion with respect to Fig.~\ref{fig:intro}(b).

Motivated by the work in~\cite{Xu_2020_CVPR}, we propose to leverage the representational capability of graphs to encode snippet-snippet, token-snippet, and token-token relations as graph edge connections.
As such, we design a new architecture referred to as Video-Language Graph Matching Network (VLG-Net)
which employs Graph Convolutional Networks (GCN)~\cite{kipf2016semi}. 
First, representation graphs for both video and language are constructed. The video graph models each snippet as a node and takes advantage of two sets of edges to represent both local temporal relations and non-local semantic relations between video snippets. Similarly, we construct a language graph, where each node is a token, and each edge reflects token-to-token relations, \eg syntactic dependencies~\cite{manning2014stanford, marcheggiani_titov_2017_encoding}. These modality-specific graphs are used to model local and non-local intra-modality context through graph convolutions. This sets the stage for addressing modality alignment by recasting inter-modality interactions as an algorithmic graph matching problem. Inspired by~\cite{li2019graph, xu2019cross}, we adopt a cross-graph attention-based matching mechanism to enable the mutual exchange of information between modalities, allowing for fine-grained alignment through a specialized set of learnable edges. Unlike some methods that focus on relatively coarse query-moment or query-snippet interactions, and similar to~\cite{chen2018temporally, zhang2019cross}, our method performs the matching operation at the level of snippets and tokens, as depicted in Fig.~\ref{fig:intro}(d). 
With this design, we avoid the need for heuristics of context modeling while learning a successful strategy for multi-modal fusion.

\noindent\textbf{Contributions.} 
\textbf{(1)} We propose VLG-Net, a new deep learning pipeline that consistently adopts graph representation for modeling modality interaction and multi-modal fusion. We address the modality fusion problem by resorting to a graph-matching approach that learns snippet-token connectivity. 
\textbf{(2)} Through extensive experiments, VLG-Net demonstrates its effectiveness in capturing modality interactions by achieving performance on par or better than state-of-the-art on three standard datasets, showing significant improvements over previously published methods in TACoS~\cite{TACoS_ACL_2013} and DiDeMo~\cite{Hendricks_2017_ICCV} datasets.

\section{Related work}\label{sec: related}

\begin{figure*}[t!]
    \centering
        \includegraphics[trim={0cm 0cm 0cm 0cm},width=15cm,clip]{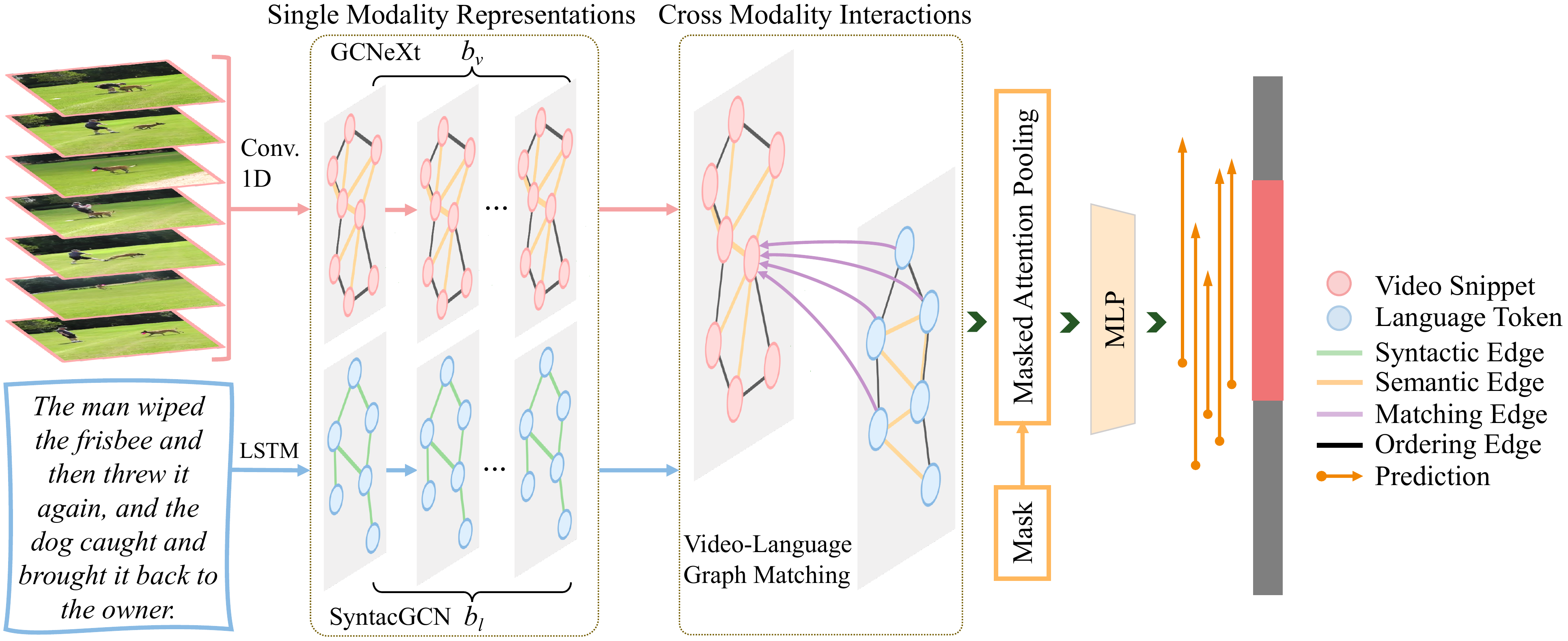} 
       \caption{\textbf{VLG-Net Architecture}. 
    Inputs are snippets features and tokens embeddings. The video stream comprises: 1D convolutions and $b_v$ GCNeXt operations. Correspondingly, tokens are fed to a stack of LSTM layers and $b_l$ SyntacGCN layers. A graph matching layer is adopted for cross-modal context modeling and multi-modal fusion. Masked attention pooling lists all possible moment candidates, and a Multi-Layer Perceptron (MLP) computes each moment's score to rank them as final predictions.}
    \label{fig:arch}
    \vspace{-0.3cm}
\end{figure*}

\subsection{Video Grounding}
\noindent \textbf{Moment candidates.} Previous works can be categorized into  proposal-free and proposal-based methods. Proposal-free approaches~\cite{chenhierarchical, Liu_2018_ECCV, Mun_2020_CVPR, Rodriguez_2020_WACV, chen2020learning, ABLR, Zeng_2020_CVPR} aim at directly regressing the temporal boundaries of the queried moment from the multi-modal fused feature. In contrast, proposal-based methods adopt a propose-and-rank pipeline by first generating moment proposals and then ranking them according to their similarity with the  textual query~\cite{Hendricks_2017_ICCV, chen_etal_2018_temporally, QSPN,10.1145/3240508.3240549, wang2020temporally,  2DTAN_2020_AAAI}. Similar to these approaches, VLG-Net is a proposal based approach.

\noindent \textbf{Moments in context.} 
For moment context modeling, some methods~\cite{ chen_etal_2018_temporally, ghosh_etal_2019_excl} attempt to use the memory property of LSTM cells~\cite{HochSchm97} to contextualize the video features. 
Alternatively, attention-based mechanisms~\cite{NIPS2017_7181} adopted in~\cite{lin2020moment,ACRN_SIGIR_18, wang2020temporally} can improve the aggregation of long-range semantic dependencies. Similar to~\cite{wang2020temporally}, we argue that visual context modeling should be dynamic and query-dependent. \cite{ 2DTAN_2020_AAAI} claims that neighbouring proposals hold valuable context and thus apply 2D convolutions (with large kernel size) to moment representations to gather context information in the latter stages of their pipeline. 
Compared to~\cite{2DTAN_2020_AAAI}, we delegate context gathering to earlier stages of our pipeline and only use a Multi-Layer Perceptron (MLP) network for moment score computation, reducing the overall computation.

\noindent \textbf{Multi-modal fusion.}
Moving beyond the simple scheme adopted in~\cite{Gao_2017_ICCV},
the work of~\cite{ijcai2018-143} devises a new cross-modality interaction scheme based on circular matrices. In~\cite{chen_etal_2018_temporally, 10.1145/3240508.3240549, wang2020temporally}, frame features are concatenated with frame-guided attention-pooled features from the query. 
Lu~\etal~\cite{lu_etal_2019_debug} take advantage of the QANet~\cite{YuDLZ00L18} architecture, which is based on cross-attention and convolution operations, for multi-modal fusion. Dynamic filters generated from language features are used in~\cite{Rodriguez_2020_WACV,Zhang_2019_CVPR} in order to modulate (through convolutions) the visual information based on the query content.
Recently, the Hadamard product has become a popular way to fuse/gate multi-modal information~\cite{Mun_2020_CVPR, Zeng_2020_CVPR, 2DTAN_2020_AAAI}.
In contrast to these methods, our graph matching layer specifically models local, non-local, and query-guided context, thereby exploiting the semantic neighbourhood of snippets and tokens to fuse the modalities through graph convolutions. Concurrently to our method, \cite{liu2020jointly} adopted attention based cross-modal graphs for fusing the video and language modality. However, opposed to our formulation,~\cite{liu2020jointly} lacks a formal design for the graph edges.

\subsection{Graphs and Graph Neural Networks}
\noindent \textbf{Graphs in Videos.} In various video understanding tasks, such as action recognition~\cite{chen2019graph, liu2019learning, wang2018videos} and action localization~\cite{Xu_2020_CVPR,zeng2019graph}, graphs can offer extensive representational power to data sequences. For example, a video can be represented as a space-time region graph~\cite{wang2018videos} or as a 3D point cloud in the spatial-temporal space~\cite{liu2019learning}. Moreover, Zeng~\etal~\cite{zeng2019graph} define temporal action proposals as nodes to form a graph, while Xu~\etal~\cite{Xu_2020_CVPR} consider video snippets as the graph nodes. Inspired by~\cite{Xu_2020_CVPR}, in VLG-Net, video snippets are represented as nodes in a graph and different specifically designed edges model their relationships.

\noindent \textbf{Graphs in Language.} 
In natural language processing (NLP), both sequential and non-local relations are crucial. The former is usually captured by recurrent neural networks~\cite{rumelhart1985learning}, while the latter can be represented using graph neural networks~\cite{ bastings2017graph, beck2018graph, marcheggiani2017encoding, 10.1007/978-3-030-00767-6_32}. 
Moreover, syntactic information has proven useful for language modeling when combined with GCNs~\cite{huang2020aligned, lin2020moment, marcheggiani_titov_2017_encoding}. 
Driven by these findings, we use LSTMs and Syntactic Graph Convolution Networks (SyntacGCN) together to model and enrich the language features in the query.

\noindent \textbf{Graph Neural Networks in Graph Matching.} 
Graph matching is one of the core problems in graph pattern recognition, aiming to find node correspondences between different graphs~\cite{caetano2009learning}. Given the ability of Graph Neural Networks (GNNs) to encode graph structure information, approaches leveraging GNNs have recently surfaced to address the graph matching problem~\cite{wang2019learning, xu2019cross}. For example, Li~\etal~\cite{li2019graph} propose a GNN-based graph matching network to represent graphs as feature vectors, which simplifies measuring their similarity. Following~\cite{li2019graph}, a neighborhood matching network is introduced by~\cite{wu2020neighborhood} to match graph nodes by estimating similarities of their neighborhoods. Due to their superiority in finding consistent correspondences between sets of features, graph matching methods have been widely applied in various tasks~\cite{jing2020visual, liu2020learning, wang2020cross, wu2020neighborhood, xu2019cross}.
Motivated by these works, we apply graph matching to the video grounding task by specifically employing a cross-graph attention matching mechanism.

\section{Methodology}\label{sec: method}
\subsection{Problem Formulation}
Given an untrimmed video and a language query, the video grounding task aims to localize a temporal moment in the video that matches the query. Each video-query pair has one associated ground-truth moment, defined as a temporal interval with boundary $(\tau_s, \tau_e)$. Our method scores $m$ candidate moments, where the $k$-th moment consists of start time $t_{s,k}$, end time $t_{e,k}$, and confidence score $p_k$. 
The video stream is represented as a sequence of $n_v$ \textbf{snippets} $V=\{v_i\}_{i=1}^{n_v}$, where each snippet has $\epsilon$ consecutive frames. Similarly, a language query is represented by $n_l$ \textbf{tokens} $L=\{l_i\}_{i=1}^{n_l}$. 
The inputs to VLG-Net are $n_v$ snippet features $X_v\in \mathbb{R}^{c_v\times n_v}$ and $n_l$ token features $X_l \in \mathbb{R}^{c_l\times n_l}$ extracted using pre-trained models, where $c_v$ and $c_l$ are the snippet and token feature dimensions. We describe the details of feature extraction in Sec.~\ref{subsec: Impl}.

\subsection{VLG-Net Architecture}
Our video grounding architecture is illustrated in Fig.~\ref{fig:arch}. 
First, we feed both the video features $X_v$ and the query embeddings $X_l$ into a stack of computation blocks. On the video path, we use 1D convolutions and GCNeXt~\cite{Xu_2020_CVPR} blocks to enrich the visual representation with local and non-local intra-modality context. On the language path, we apply LSTM and SyntacGCN~\cite{huang2020aligned} to aggregate temporal and syntactic context, which models the grammatical structure of the language query. The two paths converge in the graph matching layer for cross-modal context modeling and multi-modal fusion. After the graph matching layer, we apply masked moment attention pooling to produce the representations of possible moment candidates. Finally, we use an MLP to score the query-moment pair based on their representation and post-process the score through non-maximum suppression (NMS). We report top-$\kappa$ ranked moments as the final predictions.

\subsection{Video and Language Representations}\label{subsec: GCNeXt}
Here, we detail the set of operations performed on each modality. The stack of computation blocks of each path is specifically designed to model intra-modality context to enrich the snippet and token features.

\noindent\textbf{Video Representation.} We add 1D positional encoding, as formulated in~\cite{devlin2018bert}, to each input visual feature and apply 1D convolutions to map them to a desired dimension.
The video is then cast as a graph, where each node represents a snippet and each edge represents a dependency between a snippet pair. We design two types of edges: (i)~\textit{Ordering Edges} and (ii)~\textit{Semantic Edges}. Static \textit{Ordering Edges} connect consecutive snippets and model the temporal order. Conversely, \textit{Semantic Edges} are dynamically constructed, using the k-nearest neighbors algorithm. They connect semantically similar snippets based on their current feature representations. Specifically, an ordering or semantic snippet neighborhood is determined, and its aggregated representation is computed through edge convolutions $\F$, similar to~\cite{wang2018dynamic}. Each edge convolution employs a split-transform-merge strategy~\cite{xie2017aggregated} to increase the diversity of transformations. These graph operations (called GCNeXt) were proposed in~\cite{Xu_2020_CVPR} to enrich video snippet representations for the purpose of temporal action localization. In our architecture, we stack $b_v$ GCNeXt blocks together and refer to the input of each block as $X_v^{(i)}$ such that
\begin{align} \label{eq:gcnext}
   &X_v^{(i+1)}=\text{GCNeXt}(X_v^{(i)})= \\\nonumber
   &\sigma\left(\F{(X_v^{(i)},\mathcal{A}_o^{(i)},W_o^{(i)})}+\F{(X_v^{(i)},\mathcal{A}_s^{(i)},W_s^{(i)})}+X_v^{(i)} \right), 
\end{align}
where $X_v^{(0)}$ is the output of the convolutional layer, $\mathcal{A}_o^{(i)}$  and $\mathcal{A}_s^{(i)}$ are the adjacency matrices of~\textit{Ordering Edges} and~\textit{Semantic Edges}, respectively, and $W_{o}^{(i)}$ and $W_{s}^{(i)}$ are the trainable weights for the $i$-th GCNeXt block. 
We use Rectified Linear Unit (ReLU) as the activation function $\sigma$. Refer to~\cite{Xu_2020_CVPR} for additional details about GCNeXt. The output of the last block is referred to as $X_v^{(b_v)}$, which is the input to the graph matching layer. 

\noindent\textbf{Language Representation.} The query token features $X_l$ are fed through an LSTM of $b_s$ layers to capture semantic information and the sequential
context. Moreover, given that language follows a predefined set of grammatical rules, we set out to leverage syntactic information~\cite{marcheggiani_titov_2017_encoding, zhang_etal_2018_graph} to model grammatical inter-word relations. For this purpose, we adopt SyntacGCN, as shown in Fig.~\ref{fig:arch}. Syntactic graphs are preferred over fully connected graphs, since the former's sparsity property offers more robustness against noise in language~\cite{huang2020aligned}. Our SyntacGCN represents a query as a sparse directed graph, in which each output of the last LSTM layer, referred to as $X_l^{(0)}$, is viewed as a node, and each syntactic relation as an edge. The adjacency matrix $\mathcal{A}_l$ is directly constructed from the query's syntactic dependencies~\cite{manning2014stanford} and the graph convolution is formulated as: 
\begin{align} \label{eq:syntacGCN}
 X_{l,j}^{(i+1)} = \sigma \left(X_{l,j}^{(i)} + \sum_{k\in \mathcal{N}(j)} {\alpha_{jk}^{(i)} \mathcal{A}_{l,jk} W_l^{(i)} X_{l,k}^{(i)} }\right),
\end{align}
where $X_{l,j}^{(i)}$ is the $j$-th token feature of previous layer's output, $\mathcal{N}(j)$ is the syntactic neighbourhood of node $j$, $W_l^{(i)}$ is the learnable weight in the $i$-th layer, and $\sigma$ is ReLU. Moreover, $\alpha_{jk}^{(i)}$ is the edge weight learned from the feature of paired nodes $X_{l,j}^{(i)}$ and $X_{l,k}^{(i)}$ , defined as: 
\begin{align} \label{eq:alpha_relation}
  \alpha_{jk}^{(i)}= \mathrm{SoftMax}( \mathbf{w}_\alpha^{(i)\top} \sigma(W_\alpha^{(i)}(X_{l,j}^{(i)}||X_{l,k}^{(i)}))),
\end{align}
where $\mathbf{w}_\alpha$ and $W_\alpha$ are learnable parameters and $||$ denotes vector concatenation. We refer to the last output of the SyntacGCN as $X_l^{(b_l)}$, which will be used to match the video representation $X_v^{(b_v)}$.


\subsection{Video-Language Graph Matching}\label{subsec: Match}
\begin{figure}[!t]
    \centering
    \includegraphics[trim={3.7cm 3.7cm 1.4cm 2.25cm},width=7cm,clip]{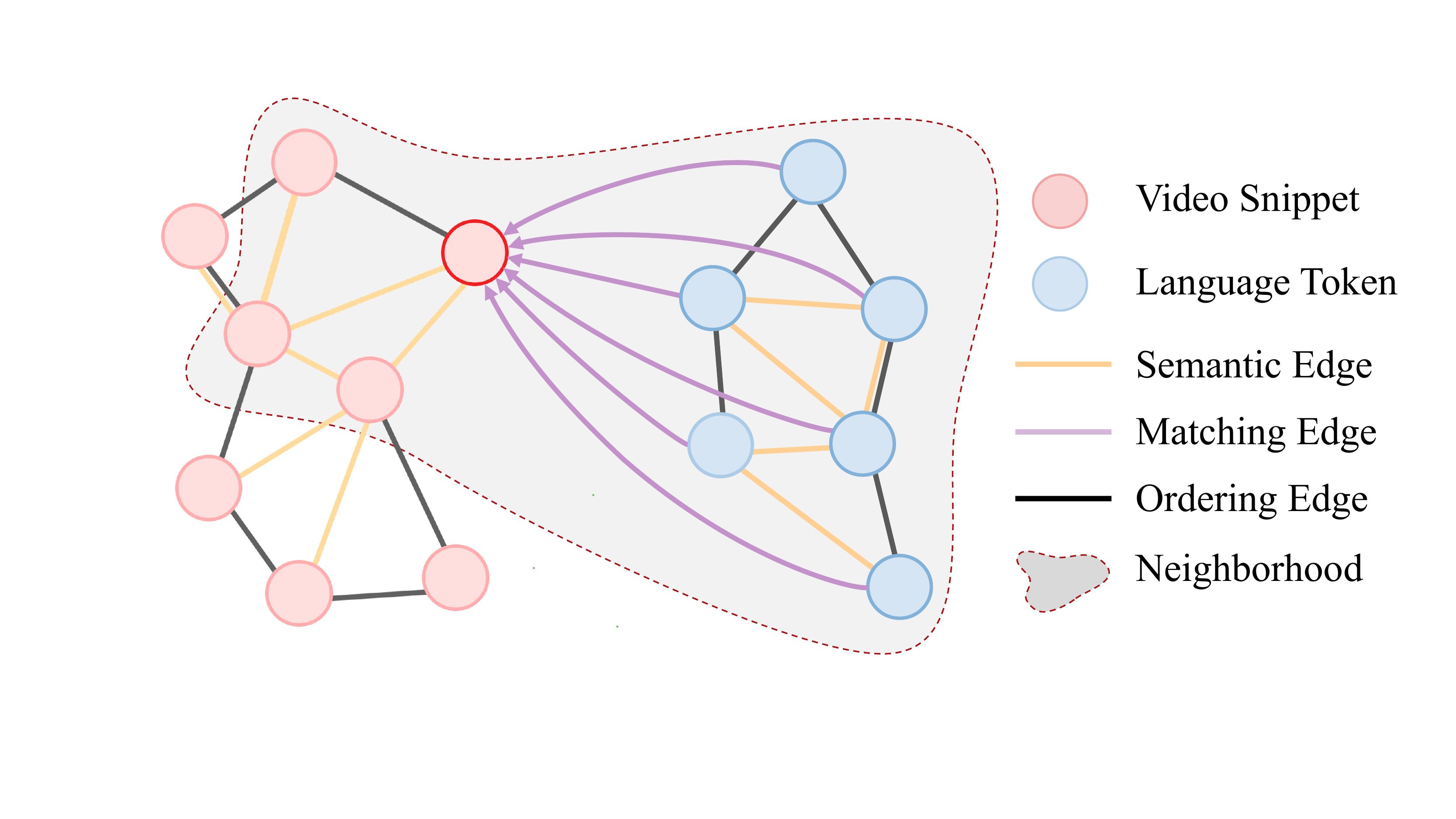}
    \caption{\textbf{The video-language matching graph.} 
    The nodes represent video snippets and query tokens. 
    Ordering Edge models the sequential nature of both modalities.
    Semantic Edge connects graph nodes in the same modality according to their feature similarity.
    Matching Edge captures the cross-modality relationships.
    We apply graph convolution on the video-language graph for cross-modal context modeling and multi-modal fusion. The neighborhood is specific for the node highlighted in red. 
    } 
    \label{fig:gmn}
\end{figure}

The enriched video representation $X_v^{(b_v)}$ and query representation $X_l^{(b_l)}$ meet and interact in the graph matching layer. This layer models the cross-modal context and allows for multi-modal fusion. 
It does so by evaluating the intra-modality correlation between each video snippet and between each query token and the inter-modality correlation between each snippet-token pair. 
The process is shown in Fig.~\ref{fig:gmn}. 
First, we create a video-language matching graph, where each node can be either a video snippet or a query token. We include three types of edges: (i)~\textit{Ordering Edge ($\mathcal{O}$)}, (ii)~\textit{Semantic Edge ($\mathcal{S}$)}, and (iii)~\textit{Matching Edge ($\mathcal{M}$)}. 

As depicted in Fig.~\ref{fig:gmn}, we use \textit{Ordering Edge} and \textit{Semantic Edge} in the video-language matching graph (as defined in Sec.~\ref{subsec: GCNeXt}), while the \textit{Matching Edge} reflects the inter-modality relation. 
\textit{Ordering Edge} models the sequential nature of both modalities. For example, if an \textit{Ordering Edge} links two tokens, the words corresponding to the two tokens are consecutive in the input query. 
\textit{Semantic Edge} is used to connect graph nodes in the same modality according to their feature similarity, providing non-local dependencies over the entire graph. 
Importantly, \textit{Matching Edge} is employed to explicitly model and learn the cross-modality interaction, to extract meaningful alignment information and learn an aggregation policy. 
The \textit{Matching Edge} weights are referenced as $\mathcal{B}$. We use \textit{Matching Edge} to densely connect all possible snippet-token pairs, and set the edge weight proportional to the correlation between the matched node features. 
Similar to \textit{Semantic Edges}, \textit{Matching Edges} are dynamic and evolve in the training process.

To combine all three types of edges, we employ relation graph convolution~\cite{schlichtkrull2017modeling} on the constructed video-language matching graph. Eq.~\ref{eq:gmm} shows the high level representation of the convolutions in this layer. Refer to the supplementary material for the full formulation. 
\begin{align} \label{eq:gmm} 
   \mathbf{X}^{(GM)}\! =\! \mathcal{A}_\mathcal{O} \mathbf{X} W_\mathcal{O}\! +\! \mathcal{A}_\mathcal{S} \mathcal{B} \mathbf{X}  W_\mathcal{S}\! +\! \mathcal{A}_\mathcal{M} \Gamma \mathbf{X} W_\mathcal{M}\! +\! \mathbf{X}
\end{align}
Here, $\mathbf{X} = \{{X_{v,1}^{(b_v)},\dots,X_{v,n_v}^{(b_v)},X_{l,1}^{(b_l)},\dots,X_{l,n_l}^{(b_l)}}\}$ is the feature representation of all the nodes in the video-language matching graph. 
$\mathcal{A}_{r}$ and $W_{r}$ for $r\in \{ \mathcal{O},\mathcal{S},\mathcal{M}\}$ represent the binary adjacency matrix and learnable weights for each set of edges. 
$\mathcal{B}$ and $\Gamma$ scale the adjacency matrices $\mathcal{A}_\mathcal{S}$ and $\mathcal{A}_\mathcal{M}$, respectively, such that $\beta_{i,j} \in \mathcal{B}$ and $\gamma_{i,j} \in \Gamma$ are proportional to $\mathbf{x}_i^{\top} \mathbf{x}_j$,
We stack together all video and language node features to form $X_v^{(GM)}\in \mathbb{R}^{c\times n_v}$ and $X_l^{(GM)}\in \mathbb{R}^{c\times n_l}$, and we pass them to the masked moment pooling layer.

\subsection{Masked Attention Pooling}\label{subsec: Pool}
The graph matching layer returns a new video graph and a new language graph fused with information from the other modality. Then, a masked attention pooling operation is applied to the new video graph to list the relevant sub-graph representations as candidate moments. The output of this module is denoted as $Y= [\mathbf{y}_k]_{k=1}^{m}, \mathbf{y}_k\in \mathbb{R}^{c}$, where $m$ is the number of candidate moments, and $c$ is the feature dimension of each moment. For efficiency purposes, the operation is implemented as a masked attention, allowing us to process each snippet feature only once, while computing each moment's representation.

\begin{figure}[!t]
    \centering
    \includegraphics[trim={0cm 0cm 0cm 0cm},width=7cm,clip]{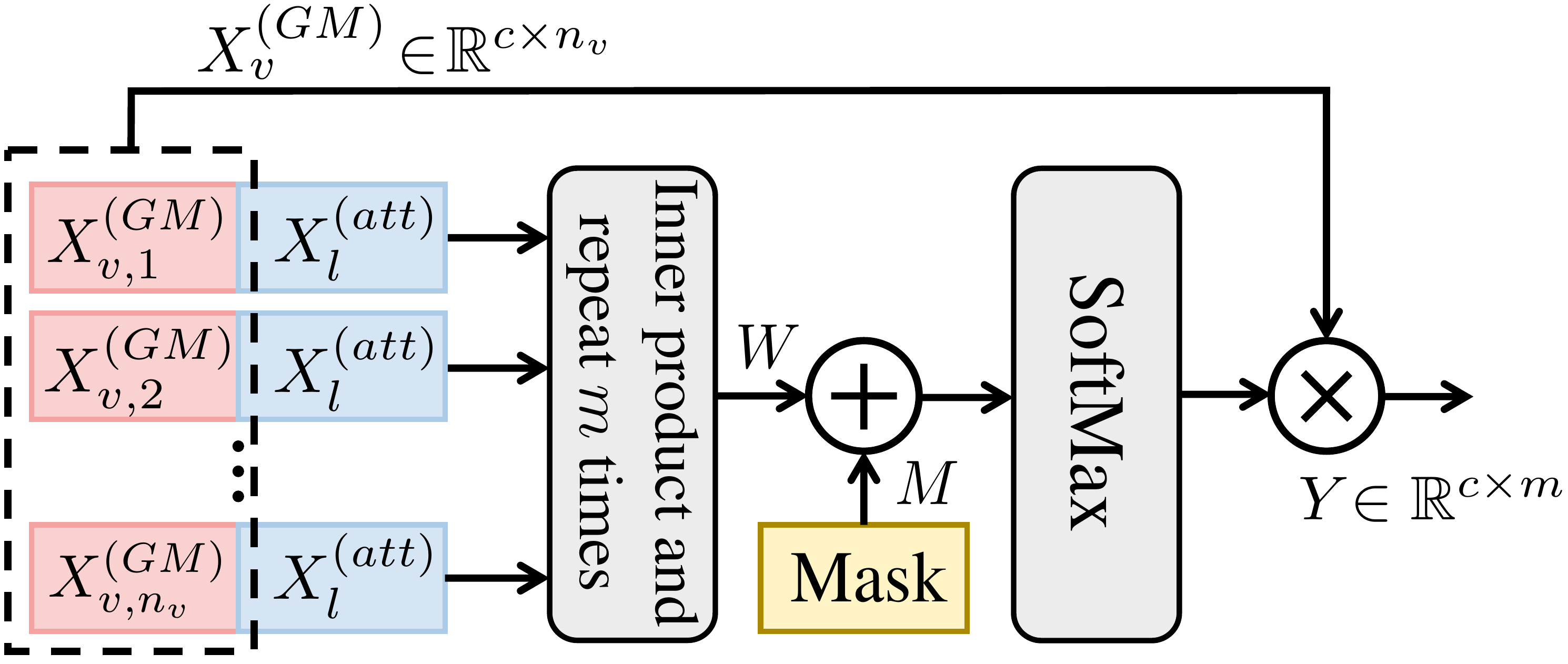}
    \caption{\textbf{Masked attention pooling.} Sequence of operations for the learnable cross-attention configuration. Inputs are video nodes $X_v^{(GM)}$ from the graph matching layer and the query embedding $X_l^{(att)}$ computed through self-attention pooling atop the graph matching output. The output $Y$ represents all moment candidates.}
    \label{fig:att}
    \vspace{-0.2cm}
\end{figure}

We implement three different schemes, namely: (i) learnable self-attention, (ii) cross-attention, and (iii) learnable cross-attention. In (i), we obtain the unnormalized attention weights by applying a 1D convolutional layer that maps each snippet feature to a single score. In (ii) and (iii), we compute the query representation $X_l^{(att)}$ by applying self-attention pooling atop the graph matching output $X_l^{(GM)}$. Cross-attention obtains the unnormalized weights by computing the inner product between the snippet and query feature, while learnable cross-attention concatenates each snippet feature with the query feature and uses a 1D convolutional layer to obtain the weights. Configuration (iii) is depicted in Fig.~\ref{fig:att}.  In all cases, the unnormalized weight vector has shape $w\in \mathbb{R}^{n_v \times 1}$ for each video. The vector is repeated $m$ times to obtain the matrix $W\in \mathbb{R}^{n_v \times m}$, and a fixed mask $M\in \mathbb{R}^{n_v \times m}$ is applied to it. Similar to  Songyang~\etal~\cite{2DTAN_2020_AAAI}, we generate $m$ moment candidates and apply a sparse sampling strategy to discard redundant moments. The value $m$ is dataset dependent. 
The mask is generated according to the sampled moments, highlighting for each of them, which are the snippets that must be taken into account when computing the moment's pooled feature. 
The attention scores are then obtained by a softmax operation. Thanks to the masking operation, snippets not related to the $n$-th moment will not be considered. Finally, the moments' features are obtained simply as a matrix multiplication $Y = X^{(GM)}  \mathrm{SoftMax}(W+M)$. 

\subsection{Moment Localization}\label{subsec: Loc}
Each candidate moment representation, output of previous module, is endowed with an additive 2D positional embedding encoding the start and end timestamps $(t_{s,k},t_{e,k}$ with $k \in [1,m])$. Then each moment feature is fed to an MLP to compute the alignment confidence score $p_k$. This score predicts the Intersection-over-Union (IoU) of each moment with the ground truth of the corresponding video-query pair. For training, we supervise this process using a cross-entropy loss, shown in Eq.~\ref{eq:loss}. Similar to~\cite{2DTAN_2020_AAAI}, we assign the label $t_k=1$ if the $IoU \geq \theta_2$, $t_k=0$ if the $IoU \leq \theta_1$, and otherwise $t_k=(IoU-\theta_1)/(\theta_2-\theta_1)$. 
\begin{equation}
    \mathcal{L} =\frac{1}{m}\sum_{k=1}^m t_k \log p_k + (1-t_k) \log(1-p_k),
\label{eq:loss}
\end{equation} 
During inference, moment candidates are ranked based on  their predicted scores and NMS is used to discard highly overlapping moments. 

\section{Experiments}\label{sec: experiments}
\subsection{Datasets} 
\noindent\textbf{ActivityNet-Captions}~\cite{Krishna_2017_ICCV} is a popular  dataset initially collected for the task of dense captioning, and recently adopted for the task of moment localization with natural language queries~\cite{chen_etal_2018_temporally,lin2020moment}. The dataset is subdivided into four splits: train, val\_1,  val\_2, and test. The test set is withheld for competition purposes, while leaving the rest publicly available. Refer to  Table ~\ref{tab:datasets}
for details about the publicly available splits. Following the setting in~\cite{lin2020moment}, we use val\_1 as the validation set and val\_2 as the testing set.

\noindent\textbf{TACoS}~\cite{TACoS_ACL_2013} consists of videos selected from the MPII Cooking Composite Activities video corpus~\cite{rohrbach2012script}. It comprises 18818 video-query pairs of different cooking activities. Each video contains an average of 148 queries, some of which are annotations of short video segments. 

\noindent\textbf{DiDeMo}~\cite{Hendricks_2017_ICCV} contains unedited video footage from Flickr with sentences aligned to unique moments in its 10642 videos. It contains more than 40k video-query pairs with coarse temporal annotations. Moment start and end points are aligned to five-second intervals and the maximum annotated moment length is 30 seconds.

Concerns regarding the Charades-STA~\cite{Gao_2017_ICCV} dataset discouraged us from evaluating our method on it. Refer to the supplementary material for a detailed discussion. 

\subsection{Implementation}\label{subsec: Impl}
\noindent\textbf{Evaluation Metrics.}
We follow the commonly used setting in ~\cite{gao2017tall}, where the Rank@$\kappa$ for IoU=$\theta$ serves as our evaluation metric. 
For example, given a video-query pair, the result is positive if any of the top-$\kappa$ predictions has IoU with the ground-truth larger or equal to $\theta$; otherwise the result is negative. We average the results across all testing samples. Following standard practice, we set $\kappa \in\{1, 5\}$ with $\theta\in\{0.3,0.5,0.7\}$ for ActivityNet Captions,
$\kappa\in\{1, 5\}$ with $\theta\in\{0.1,0.3,0.5\}$ for TACoS,  and $\kappa\in\{1, 5\}$ with $\theta\in\{0.5,0.7, 1.0\}$ for DiDeMo.

\noindent\textbf{Language and Video Features.}
After lower-case conversion and tokenization, we use the pretrained GloVe model~\cite{pennington2014glove} to obtain the initial query embedding for every token and extract the syntactic dependencies using the Stanford CoreNLP 4.0.0 parser~\cite{manning2014stanford}. The $b_s$ layers 
of LSTM with 512 hidden units are used as the query encoder.  Then, the syntactic GCN encodes syntactic information of the queries and returns a new embedding with 512 dimensions. For visual features, we use pretrained C3D~\cite{tran2015learning} for ActivityNet Captions and TACoS, and VGG16~\cite{simonyan2014very} for DiDeMo,  while holding their parameters fixed during training, as they are readily available and commonly used by state-of-the-art methods. We use 1D convolutions to project the input visual features to a fixed dimension (512), and the GCNeXt blocks' hyper-parameters are set as in~\cite{Xu_2020_CVPR}. 

\begin{table}[!t]
    \centering 
\resizebox{\linewidth}{!}{%
\scalebox{1.0}{
\begin{tabular}{l|c|ccc|c} 
\toprule
Dataset             &   Num.   & \multicolumn{3}{c|}{Video-Sentence pairs} & Vocab.         \\ 
                    &   Videos & train & val & test                       & Size    \\ 
\midrule
ActivityNet Captions~\cite{Krishna_2017_ICCV} & 14926   & 37421  & 17505  & 17031 & 15406          \\
TACoS~\cite{TACoS_ACL_2013}                   & 127     & 10146  & 4589   & 4083   & 2255           \\
DiDeMo~\cite{Hendricks_2017_ICCV}             & 10642   & 33005  & 4180   & 4021   & 7523           \\
\midrule
Charades-STA~\cite{Gao_2017_ICCV}             & 6670    & 12404  & $-$      & 3720      & 1289           \\
\bottomrule
\end{tabular}
}
}
\vspace{.1cm}
\caption{\label{tab:datasets}{\bf Datasets statistics.} We report relevant information for each datasets available for the grounding task. }

    \vspace{-0.5cm}
\end{table}

\noindent\textbf{Implementation details.}
We use Adam~\cite{kingma2014adam} with a StepLR scheduler~\cite{loshchilov2016sgdr}. 
We adopt learning rates ranging from $10^{-3}$ to $10^{-4}$ for different datasets. The number of sampled snippets $n_v$ is set to $64$ for ActivityNet Captions, and $256$ for TACoS, and $48$ for DiDeMo. The values of ($b_v$, $b_s$, $b_l$) are equal to ($1,3,4$), ($4,5,2$), ($2,3,4$), respectively for the three datasets. In post-processing, we apply NMS with values respectively to the $m$ predictions to filter out highly overlapping moments. Values of $m$ for each dataset are: $1104$, $3101$, $505$ while the NMS thresholds are: $0.5$, $0.3$, $0.5$. We adapt BCE with logits loss to make the training process more numerically stable. The IoU thresholds ($\theta_1$, $\theta_1$) for the three datasets are: ($0.7$, $0.71$), ($0.5$, $0.7$), ($0.69$, $1.0$).

\subsection{Comparison with State-of-the-Art}
Comparisons are carried out only against methods using the same input features as VLG-Net. In the Tables, we highlight the top-1 and top-2 performance values by bold and underline, respectively. 

\noindent\textbf{ActivityNet Captions (Table \ref{tab:anet}).}
VLG-Net offers the highest performance for the tight IoU=0.7. However, it falls short against~\cite{liu2020jointly} on the lousier IoU=0.5. In practical terms, the two methods are to be considered on par. Nonetheless, notice how tighter IoU translates to better retrieval quality in a real use-case scenario. In these terms, VLG-Net is to be preferred over the competitive~\cite{liu2020jointly}. Finally, notice how VLG-Net achieves a significant boost against the recently released 2D-TAN~\cite{2DTAN_2020_AAAI} and DRN~\cite{Zeng_2020_CVPR}.

\noindent\textbf{TACoS (Table~\ref{tab:tacos}).}
Our model outperforms state-of-the-art methods and achieves the highest scores for all IoU thresholds with significant improvements. In particular, VLG-Net exceeds the previous art~\cite{liu2020jointly,Zeng_2020_CVPR,2DTAN_2020_AAAI} by a large margin, ranging from $7.10$\% to $11.52$\%, across all evaluation settings, showcasing the excellent design of the architecture. 

\noindent\textbf{DiDeMo (Table~\ref{tab:didemo}).  }
Our proposed technique outperforms the top-ranked methods~\cite{liu2020jointly,ACRN_SIGIR_18,10.1145/3240508.3240549} with respect to R@1 and R@5 for IoU0.5 and 0.7 with evident increases. It also reaches the highest performance in regards to R@1 IoU1.0. For R@5 IoU1.0, VLG-Net ranks second, falling short with respect to TMN~\cite{chen_etal_2018_temporally} by $1.32\%$.
For completeness, we report the performances of TGN~\cite{Liu_2018_ECCV} and TMN~\cite{chen_etal_2018_temporally}; with the caveat that their performance could not be verified as the code is not made publicly available. Hence the different colour for the corresponding rows in Table~\ref{tab:didemo}.

\begin{table}[!t]
    \centering 
\setlength{\tabcolsep}{3pt}
\renewcommand{\arraystretch}{1} 
\resizebox{0.8\linewidth}{!}{%
\scalebox{1.0}{
\begin{tabular}{l|cc|cc} 
\toprule
&   \multicolumn{2}{c|}{R@1  }& \multicolumn{2}{c}{R@5  } \\ 
&   IoU0.5 & IoU0.7 & IoU0.5 & IoU0.7 \\ 
\midrule
MCN~\cite{Hendricks_2017_ICCV} &  $21.36$ & $6.43$ &  $53.23$ & $29.70$  \\
CTRL~\cite{Gao_2017_ICCV}  & $29.01$ & $10.34$ & $59.17$ & $37.54$  \\
TGN~\cite{chen_etal_2018_temporally}  & $27.93$ & $-$ &  $44.20$ & $-$  \\
ACRN~\cite{Liu_2018_ECCV} &  $31.67$ & $11.25$ & $60.34$ & $38.57$  \\
CMIN~\cite{lin2020moment}  & $44.62$ & $24.48$ & $69.66$ & $52.96$  \\
ABLR~\cite{ABLR}  & $36.79$ & $-$ & $-$ & $-$  \\
TripNet~\cite{hahn2020tripping} & $32.19$ & $13.93$ & $-$ & $-$  \\
PMI~\cite{chen2020learning} & 38.28 & 17.83 &$-$ & $-$  \\
2D-TAN (P)~\cite{2DTAN_2020_AAAI} & $44.51$ & $26.54$ & $77.13$ &	$61.96$ \\
2D-TAN (C)~\cite{2DTAN_2020_AAAI} & $44.05$ & $27.38$ & $76.65$ &	$\underline{62.26}$ \\
DRN~\cite{Zeng_2020_CVPR}    & $45.45$ & $24.36$ & $\mathbf{77.97}$ & $50.30$ \\
CSMGAN~\cite{liu2020jointly} & $\mathbf{49.11}$ & $\underline{29.15}$ & $\underline{77.43}$ & $59.63$ \\
\midrule
VLG-Net   & $\underline{46.32}$ & $\mathbf{29.82}$ & $77.15$ &	$\mathbf{63.33}$ \\
\bottomrule
\end{tabular}
}
}
\vspace{.1cm}
\caption{\label{tab:anet}{\bf State-of-the-art comparison on ActivityNet Captions.} We report the results at different Recall@$\kappa$ and different IoU thresholds. VLG-Net reaches the highest scores for IoU0.7 for both R@1 and R@5. 
}
\label{anet_results}

    \vspace{-0.2cm}
\end{table}

\subsection{Ablation Study} \label{subsec: Abl}
To motivate our design choices, we present two ablations that focus on relevant aspects of our method. The first ablation showcases the importance of context modeling. The second investigates VLG-Net's performance when other commonly adopted multi-modal fusion operations replace the graph matching module. In Table ~\ref{tab:ablation}, we report our best result (first column) and summarize all ablated variants. For simplicity, we evaluate on the TACoS dataset and specifically focus on its most challenging setups: R@1 IoU0.5 and R@5 IoU0.5. To promote fair comparison, we report each model's capacity in millions (M) of parameters. More ablations are reported in the supplementary material.

\noindent\textbf{Context Ablation.} First, we investigate the impact of different context modeling strategies and compare six variants with our VLG-Net. VLG-Net\textsubscript{NC} represents a ``No Context'' architecture, in which we replace the GCNeXt and SyntacGCN modules with fully connected layers that do not model any intra-modality context. Moreover, we switch off the \textit{Ordering} and \textit{Semantic Edges} in the graph matching module. 
Although the model capacity for VLG-Net\textsubscript{NC} only drops $0.21$M ($1.6\%$), its performance degrades up to $8.82\%$  with respect to VLG-Net.
Following~\cite{Hendricks_2017_ICCV}, we devise VLG-Net\textsubscript{GM} (``Global on Moments'') and VLG-Net\textsubscript{GI} (``Global on Input'') experiments. The first one extends VLG-Net\textsubscript{NC} by concatenating each moment feature with a global video feature after the matching operation. Instead, in VLG-Net\textsubscript{GI}, we concatenate each snippet and token feature with an average pooled version of their respective raw input features. 
Differently, following~\cite{Gao_2017_ICCV}, VLG-Net\textsubscript{LM} (``Local on Moments'') models local context by extending the moment's boundaries when computing the moment's features in the masked attention pooling module. 
The simple context modeling adopted in these architectures allows them to improve their performance with respect to VLG-Net\textsubscript{NC} up to $2.32\%$. Nonetheless, they fall short of VLG-Net by $6.5 - 7.4 \%$. 
Note that VLG-Net\textsubscript{GM} and VLG-Net\textsubscript{GI} have a larger model capacity, $0.07$M and $0.41$M, respectively, compared to VLG-Net.
VLG-Net\textsubscript{T} (``Temporal Context Only'') is as VLG-Net, where we remove each \textit{Semantic Edge} and replace the SyntacGCN layers with GCNeXt ones. In this variant, only temporal dependencies are modeled. VLG-Net\textsubscript{S} (``Semantic Context Only'') does not model temporal dependencies but just the semantic ones. 
Both models' performance surpasses $30 \%$ for R@1 IoU0.5, which indicates that the introduction of GCNeXt and SyntacGCN layers can boost the performance.  
Our final architecture takes advantage of both modules, achieving the best results. These ablations demonstrate that temporal and semantic context are complementary and showcase the benefits of the proposed context modeling strategy. 

\begin{table}[!t]
    \centering 
\setlength{\tabcolsep}{2pt}
\renewcommand{\arraystretch}{1} 
\resizebox{\linewidth}{!}{%
\scalebox{1.0}{
\begin{tabular}{l|ccc|ccc} 
\toprule
                                     &   \multicolumn{3}{c|}{R@1  }& \multicolumn{3}{c}{R@5  }   \\ 
                                     &  IoU0.1 & IoU0.3  & IoU0.5  & IoU0.1  & IoU0.3  & IoU0.5  \\ 
\midrule
MCN~\cite{Hendricks_2017_ICCV}       & $14.42$ & $-$     & $5.58$  & $37.35$ & $-$     & $10.33$  \\
CTRL~\cite{Gao_2017_ICCV}            & $24.32$ & $18.32$ & $13.30$ & $48.73$ & $36.69$ & $25.42$  \\
MCF~\cite{wu2018multi}               & $25.84$ & $18.64$ & $12.53$ & $52.96$ & $37.13$ & $24.73$  \\
TGN~\cite{chen_etal_2018_temporally} & $41.87$ & $21.77$ & $18.90$ & $53.40$ & $39.06$ & $31.02$  \\
ACRN~\cite{ACRN_SIGIR_18}            & $24.22$ & $19.52$ & $14.62$ & $47.42$ & $34.97$ & $24.88$  \\
ROLE~\cite{10.1145/3240508.3240549}  & $20.37$ & $15.38$ & $9.94$  & $45.45$ & $31.17$ & $20.13$  \\
VAL~\cite{song2018val}               & $25.74$ & $19.76$ & $14.74$ & $51.87$ & $38.55$ & $26.52$  \\
ACL-K~\cite{Ge_2019_WACV}            & $31.64$ & $24.17$ & $20.01$ & $57.85$ & $42.15$ & $30.66$  \\
CMIN~\cite{lin2020moment}            & $36.68$ & $27.33$ & $19.57$ & $64.93$ & $43.35$ & $28.53$  \\
SM-RL~\cite{wang2019language}        & $26.51$ & $20.25$ & $15.95$ & $50.01$ & $38.47$ & $27.84$  \\
SLTA~\cite{jiang2019cross}           & $23.13$ & $17.07$ & $11.92$ & $46.52$ & $32.90$ & $20.86$  \\
SAP~\cite{Chen_19_SAP}               & $31.15$ & $-$     & $18.24$ & $53.51$ & $-$     & $28.11$  \\
TripNet~\cite{hahn2020tripping}      & $-$     & $23.95$ & $19.17$ & $-$     & $-$     & $-$      \\
2D-TAN (P)~\cite{2DTAN_2020_AAAI}    & $\underline{47.59}$ & $\underline{37.29}$ & $25.32$ & $70.31$ & $\underline{57.81}$ & $\underline{45.04}$ \\
2D-TAN (C)~\cite{2DTAN_2020_AAAI}    & $46.44$ & $35.22$ & $25.19$ & $\underline{74.43}$ & $56.94$ &	$44.21$ \\
DRN~\cite{Zeng_2020_CVPR}            & $-$     & $-$     & $23.17$ & $-$     & $-$     & $33.36$  \\
CSMGAN~\cite{liu2020jointly}         & $42.74$ & $33.90$ & $\underline{27.09}$ & $68.97$ & $53.98$ & $41.22$ \\
\midrule
VLG-Net   &  $\mathbf{57.21}$ & $\mathbf{45.46}$ & $\mathbf{34.19}$ & $\mathbf{81.80}$ & $\mathbf{70.38}$ & $\mathbf{56.56}$ \\
\bottomrule
\end{tabular}
}
}
\vspace{.1cm}
\caption{\label{tab:tacos}{\bf State-of-the-art comparison on TACoS.}  
Our model outperforms all previous methods achieving significantly higher performance with great margins on all metrics. 
}

\end{table}

\begin{table}[!t]
    \centering 
\setlength{\tabcolsep}{3pt}
\renewcommand{\arraystretch}{1} 
\resizebox{\linewidth}{!}{%
\scalebox{1.0}{
\begin{tabular}{l|ccc|ccc} 
\toprule
&   \multicolumn{3}{c|}{R@1  }& \multicolumn{3}{c}{R@5  } \\ 
&   IoU0.5 & IoU0.7 & IoU1.0 & IoU0.5 & IoU0.7 & IoU1.0 \\ 
\midrule
MCN \cite{Hendricks_2017_ICCV} &  $-$ & $-$ &  $13.10$ & $-$ & $-$ & $44.82$ \\
\textcolor{mygray}{TMN~\cite{Liu_2018_ECCV}} &  \textcolor{mygray}{$-$} &  \textcolor{mygray}{$-$} &  \textcolor{mygray}{$18.71$} &  \textcolor{mygray}{$-$} &  \textcolor{mygray}{$-$} & \textcolor{mygray}{$\mathbf{72.97}$} \\
\textcolor{mygray}{TGN~\cite{chen_etal_2018_temporally}}  &   \textcolor{mygray}{$-$} &  \textcolor{mygray}{$-$} &  \textcolor{mygray}{$\underline{24.28}$} &  \textcolor{mygray}{$-$} &  \textcolor{mygray}{$-$} & \textcolor{mygray}{$71.43$} \\
ACRN \cite{ACRN_SIGIR_18}           &  $27.44$ & $16.65$ &  $-$ & $69.43$ & $29.45$ & $-$ \\
ROLE \cite{10.1145/3240508.3240549} &  $29.40$ & $15.68$ &  $-$ & $70.72$ & $33.08$ & $-$ \\
CSMGAN~\cite{liu2020jointly} & $\underline{29.44}$ & $\underline{19.16}$&  $-$ & $\underline{70.77}$ & $\underline{41.61}$ & $-$\\
\midrule
VLG-Net  &  $\mathbf{33.35}$ & $\mathbf{25.57}$ & $\mathbf{25.57}$ & $\mathbf{88.86}$ & $\mathbf{71.72}$ & $\underline{71.65}$\\
\bottomrule
\end{tabular}
}
}
\vspace{.1cm}
\caption{\label{tab:didemo}{\bf State-of-the-art comparison on DiDeMo.} Our proposed model outperforms the top ranked method ROLE and ACRN with respect to IoU0.5 and 0.7 for R@1 and R@5 with clear margins. It also reaches the highest performance in regards to R@1 IoU1.0. }

    \vspace{-0.2cm}
\end{table}

\begin{figure*}[!ht]
    \centering
    \includegraphics[trim={0cm 0cm 0cm 0cm},width=\linewidth,clip]{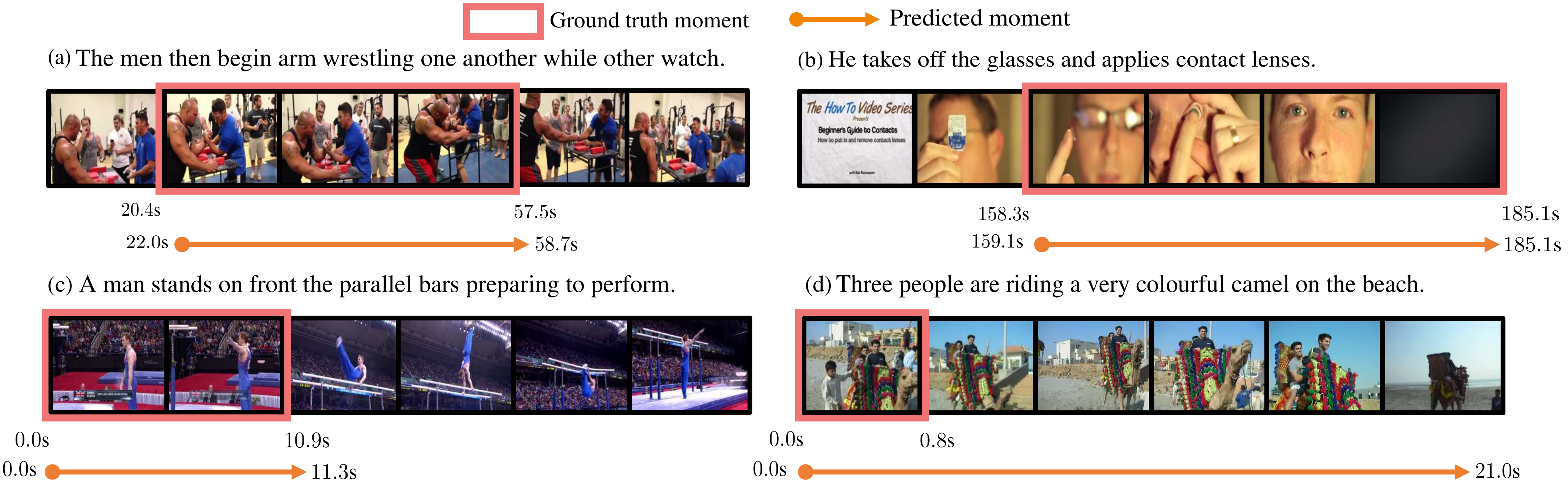}
    \caption{\textbf{Qualitative Results.} Examples of grounding results, we compare ground truth annotations (box) and predicted temporal endpoints (arrow). See Section~\ref{subsec: qualitative} for more details.} 
    \label{fig:vis}
\end{figure*}

\begin{table*}[!ht]
    \centering 
\setlength{\tabcolsep}{2pt}
\renewcommand{\arraystretch}{1.1} 
\resizebox{\textwidth}{!}{%
\scalebox{1.0}{
\begin{tabular}{l|c|cccccc|ccc} 
\toprule
& \multirow{2}*{\textbf{VLG-Net}} &  \multicolumn{6}{c|}{Context Ablation } &  \multicolumn{3}{c}{Multi-modal Fusion Ablation }   \\
      &  & VLG-Net\textsubscript{NC} &  VLG-Net\textsubscript{GM} &  VLG-Net\textsubscript{LM} &  VLG-Net\textsubscript{GI} &  VLG-Net\textsubscript{T} &  VLG-Net\textsubscript{S} &  VLG-Net\textsubscript{HM} &  VLG-Net\textsubscript{HS} &  VLG-Net\textsubscript{CS}   \\
\midrule
R@1 IoU0.5 & $\mathbf{34.19}$ & $25.37$& $27.12$  & $26.82$  &  $27.69$ & $30.34$  & $30.94$  &  $27.27$ & $27.54$  & $28.39$  \\
R@5 IoU0.5 & $\mathbf{56.56}$ & $48.24$& $48.56$  & $47.11$  &  $49.39$ & $51.84$  & $53.14$  &  $49.99$ & $48.54$  & $51.31$  \\
\# of Parameters (M) & $\mathbf{13.29}$ & $13.08$ &  $13.35$ &  $13.08$  &  $13.70$ &  $13.22$  &  $13.22$  &  $13.28$  &  $13.28$  &  $13.28$  \\
\bottomrule                    
\end{tabular}
}
}
\vspace{.1cm}
\caption{\label{tab:ablation}{\bf Ablation on context modelling and multi-modal fusion approaches.} This ablation shows how the pipeline takes advantage of context modelling and our graph matching module to achieving state-of-the-art performance.}

    \vspace{-0.2cm}
\end{table*}

\noindent\textbf{Multi-modal Fusion Ablation.} To evaluate our graph matching module's capabilities for multi-modality fusion, we replace it with other commonly used operations in the literature. Following~\cite{2DTAN_2020_AAAI}, we use a Hadamard product between the video moment's features and an average pooled feature for the language query. We adopt several fully connected layers before and after the fusion to keep the model size close to ours. As compared to~\cite{2DTAN_2020_AAAI}, we adopt the learnable masked attention pooling for generating the moment's features, which allows for interactions between query tokens and video snippets before the fusion operation. We refer to this model as VLG-Net\textsubscript{HM} (``Hadamard on Moments'').
We also apply the Hadamard product at the snippet level~\cite{Mun_2020_CVPR,Rodriguez_2020_WACV,Zeng_2020_CVPR} and train VLG-Net\textsubscript{HS} (``Hadamard on Snippets''). Finally, following~\cite{chen_etal_2018_temporally,chenhierarchical,wang2020temporally}, we concatenate each snippet feature with the query feature and use linear layers for the projection in VLG-Net\textsubscript{CS} (``Concatenate on Snippets''). 
We can conclude that fusing the modalities at the snippet level tends to perform better. The Hadamard operation has gained quite some traction for its good performance and absence of trainable parameters, making it efficient to compute. However, we argue that the complexity of multi-modal alignment calls for more elaborate strategies for multi-modal fusion. Our graph matching module offers a perspective in this research direction. 

\subsection{Visualization} \label{subsec: qualitative}
We show several qualitative grounding results from ActivityNet Captions in Fig.~\ref{fig:vis}. Our VLG-Net can generate precise moment boundaries that match the query well in different scenarios. Worth mentioning, our method can sometimes give predictions that are more meaningful than the ground truth annotation. As shown in Fig.~\ref{fig:vis}(d), although the ground truth only aligns to the beginning of the video, the query ``\textit{Three people are riding a very colourful camel on the beach.}'' can semantically match the whole video. In this case, our VLG-Net gives a more reasonable grounding result. Additional visualizations are reported in the supplementary material. 

\section{Conclusion}\label{sec: conclusions}
This paper addresses the problem of text-to-video temporal grounding, where we cast the problem as an algorithmic graph matching.
We propose Video-Language Graph Matching Network (VLG-Net) to match the video and language modalities. We represent each modality as graphs and explore four types of edges, \textit{Syntactic Edge}, \textit{Ordering Edge}, \textit{Semantic Edge}, and \textit{Matching Edge}, to encode local, non-local, and cross-modality relationships to align the video-query pair. Extensive experiments show that our VLG-Net can model inter- and intra-modality context, learn multi-modal fusion and surpass the current state-of-the-art performance on three widely used datasets. 

\noindent\textbf{Acknowledgments.} This work was supported by the King Abdullah University of Science and Technology (KAUST) Office of Sponsored Research through the Visual Computing Center (VCC) funding.

{\small
\bibliographystyle{ieee_fullname}
\bibliography{egbib}
}

\newpage
\section*{Supplementary Material}
\subsection*{Formulation of Video-Language Graph Matching}
In this section, we provide a detailed overview and formulation of the video-language graph matching. 
This inputs to this layer are the enriched video representation $X^{(bv)}_v$ and query representation $X^{(bl)}_l$ outputs of the single modality stack of computational blocks. The graph matching layer models the cross-modal context and allows for multi-modal fusion. To this purpose the video-language matching graph is constructed and three types of edges are designed: (i) \textit{Ordering Edge} ($\mathcal{O}$), (ii) \textit{Semantic Edge} ($\mathcal{S}$), and (iii) \textit{Matching Edge }($\mathcal{M}$).

To aggregate the information, we employ relation graph convolution~\cite{schlichtkrull2017modeling} on the constructed video-language matching graph. Eq.~\ref{eq:gmm} shows the high level representation of the convolutions in this layer.
\begin{align} \label{eq:gmm} 
   \mathbf{X}^{(GM)}\! =\! \mathcal{A}_\mathcal{O} \mathbf{X} W_\mathcal{O}\! +\! \mathcal{A}_\mathcal{S} \mathcal{B} \mathbf{X}  W_\mathcal{S}\! +\! \mathcal{A}_\mathcal{M} \Gamma \mathbf{X} W_\mathcal{M}\! +\! \mathbf{X}
\end{align}
Here, $\mathbf{X} = \{{X_{v,1}^{(b_v)},\dots,X_{v,n_v}^{(b_v)},X_{l,1}^{(b_l)},\dots,X_{l,n_l}^{(b_l)}}\}$ is the feature representation of all the nodes in the video-language matching graph. 
$\mathcal{A}_{r}$ and $W_{r}$ for $r\in \{ \mathcal{O},\mathcal{S},\mathcal{M}\}$ represent the binary adjacency matrix and learnable weights for each set of edges. 
Specifically, $\mathcal{B}$ and $\Gamma$ scale the adjacency matrices $\mathcal{A}_\mathcal{S}$ and $\mathcal{A}_\mathcal{M}$. Both $\beta_{i,j}\in \mathcal{B}$ and $\gamma_{i,j} \in \Gamma$ are proportional to $\mathbf{x}_i^{\top} \mathbf{x}_j$,
\begin{align}
      \beta_{i,j}= \frac{\exp{[\mathbf{x}_i^{\top} \mathbf{x}_j}]} 
      {\sum_{\mathcal{A}_\mathcal{S}(k,j)=1} \exp{[\mathbf{x}_k^{\top} \mathbf{x}_j}]}, \\
      \gamma_{i,j}= \frac{\exp{[\mathbf{x}_i^{\top} \mathbf{x}_j}]} 
      {\sum_{\mathcal{A}_\mathcal{M}(k,j)=1} \exp{[\mathbf{x}_k^{\top} \mathbf{x}_j}]}.
\end{align}

In practise, to implement GPU-memory efficient graph convolution operation, we replace the time-consuming matrix multiplication by indexing operation of tensors. 
Thus, the semantic and matching edge convolution can be present as 
\begin{align} \label{eq:gmm2} 
   \mathcal{A}_\mathcal{S} \mathcal{B} \mathbf{X}  W_\mathcal{S} = \sum_{j\in \mathcal{N}_i^\mathcal{S}} ( \hat{W}_\mathcal{S}^T [\beta_j\mathbf{x}_{j}||\mathbf{x}_{i}]), \\
   \mathcal{A}_\mathcal{M} \Gamma \mathbf{X} W_\mathcal{M} = \sum_{j\in \mathcal{N}_i^\mathcal{M}} ( \hat{W}_\mathcal{M}^T [\gamma_j\mathbf{x}_{j}||\mathbf{x}_{i}]),
\end{align}
where $\mathcal{N}_i^\mathcal{*}$ is the neighbourhood of node $i$ connected by edge with type $*$, $* \in \{\mathcal{S},\mathcal{M}\}$. The $||$ sign means concatenation of features. $\hat{W}_\mathcal{S}, \hat{W}_\mathcal{M}$ are learnable weights.

Moreover, as shown by A.2 of G-TAD\cite{Xu_2020_CVPR}, our ordering edge convolution, can be efficiently computed as a 1D convolution with kernel size 3.
\begin{align} \label{eq:gmm3} 
   \mathcal{A}_\mathcal{O} \mathbf{X} W_\mathcal{O} = Conv1D[X]
\end{align}
Therefore, we can equivalently formulate Eq.~\ref{eq:gmm} as:
\begin{align} \label{eq:gmm_final} 
\begin{split}
   \mathbf{X}^{(GM)} & =  Conv1D[X] \\ 
   & + \sum_{j\in \mathcal{N}_i^\mathcal{S}} ( \hat{W}_\mathcal{S}^T [\beta_j\mathbf{x}_{j}||\mathbf{x}_{i}]) \\ 
   & + \sum_{j\in \mathcal{N}_i^\mathcal{M}} ( \hat{W}_\mathcal{M}^T [\gamma_j\mathbf{x}_{j}||\mathbf{x}_{i}]) \\
   & + \mathbf{X}
\end{split}
\end{align}

\subsection*{Graph matching edges ablation}
We ablate the contribution of the three different types of edges designed for the graph matching module. 
We report in Table~\ref{tab:ablation2}  the performance of VLG-Net for the TACoS dataset when each edge is removed from the architecture.
As previously stated, the \textit{Ordering Edges} or \textit{Semantic Edges} are responsible for aggregating contextual information within the graph matching module. When removed, they lead to noticeable degradation of the performance of $2.15 \%$ and $3.77 \%$, respectively.
Conversely, as expected, when the \textit{Matching Edges} are removed, the performances are severely impaired. We assist in a drop of $27.34 \%$, showcasing the high relevance of the matching operation. 
Note that, the removal of the \textit{Matching Edges} prevents the fusion between the modalities. Nonetheless, the two modalities still interact in the Masked Attention Pooling module through the learnable cross-attention pooling method. However, this limited interaction cannot bridge the complex semantic information between modalities. 
The ablation showcases the importance of designing effective operation for multi-modal fusion to achieve high performance on the grounding task. 
Nonetheless, we can conclude that all edges are relevant and necessary to obtain state-of-the-art performance.

\begin{table}[!h]
    \centering 
\resizebox{0.85\linewidth}{!}{%
\scalebox{1.0}{
\begin{tabular}{l|ccc|c} 

\toprule
 
\multirow{2}{*}{Dataset} & \multicolumn{3}{c|}{Edge Types} & \multirow{2}{*}{R@1 IoU0.5} \\
& Ordering & Semantic & Matching \\ 
\midrule
\multirow{4}{*}{TACoS}  & \cmark & \cmark & \cmark &  $\mathbf{34.19}$ \\
                        & \xmark & \cmark & \cmark &  $32.04$          \\
                        & \cmark & \xmark & \cmark &  $30.42$          \\
                        & \cmark & \cmark & \xmark &  $6.85$          \\
\bottomrule
\end{tabular}
}
}
\vspace{.1cm}
\caption{\label{tab:ablation2}{\bf Ablation of different edges.} 
We investigate the impact of edges within the graph matching layer. We report the performance of our VLG-Net when specific edges are removed, as well as our best performance for TACoS datasets. }
\end{table}

\subsection*{Visualization graph matching attention}

\begin{figure}[!ht]
\captionsetup[subfigure]{labelformat=empty}
\begin{center}

    \subfloat[(a) Query: ``She is holding an accordian as she talks.'']{\includegraphics[width=0.75\linewidth,clip]{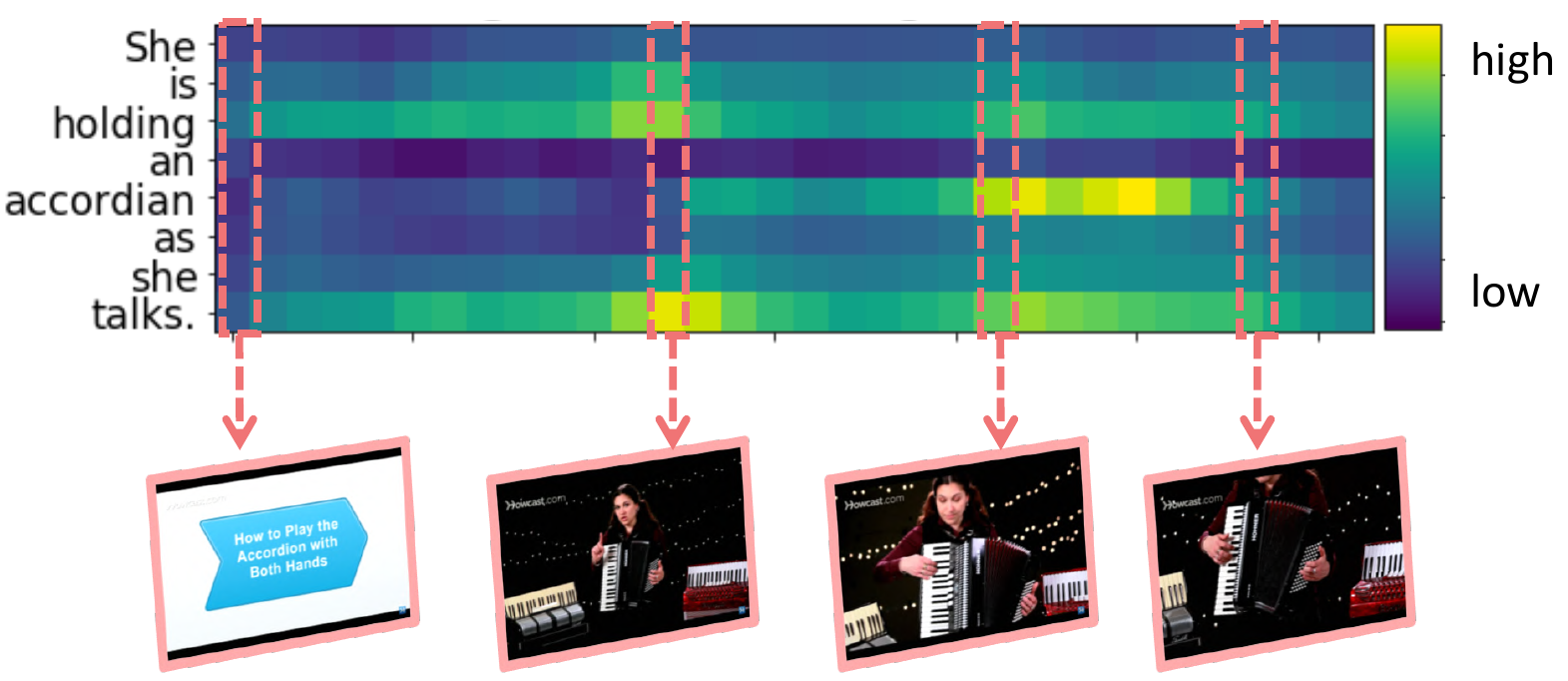} \label{fig:supp_att_vis_a}}
    
    \subfloat[\centering (b) Query: ``Three people are riding a\linebreak very colorful camel on the beach.'']{\includegraphics[width=0.75\linewidth,clip]{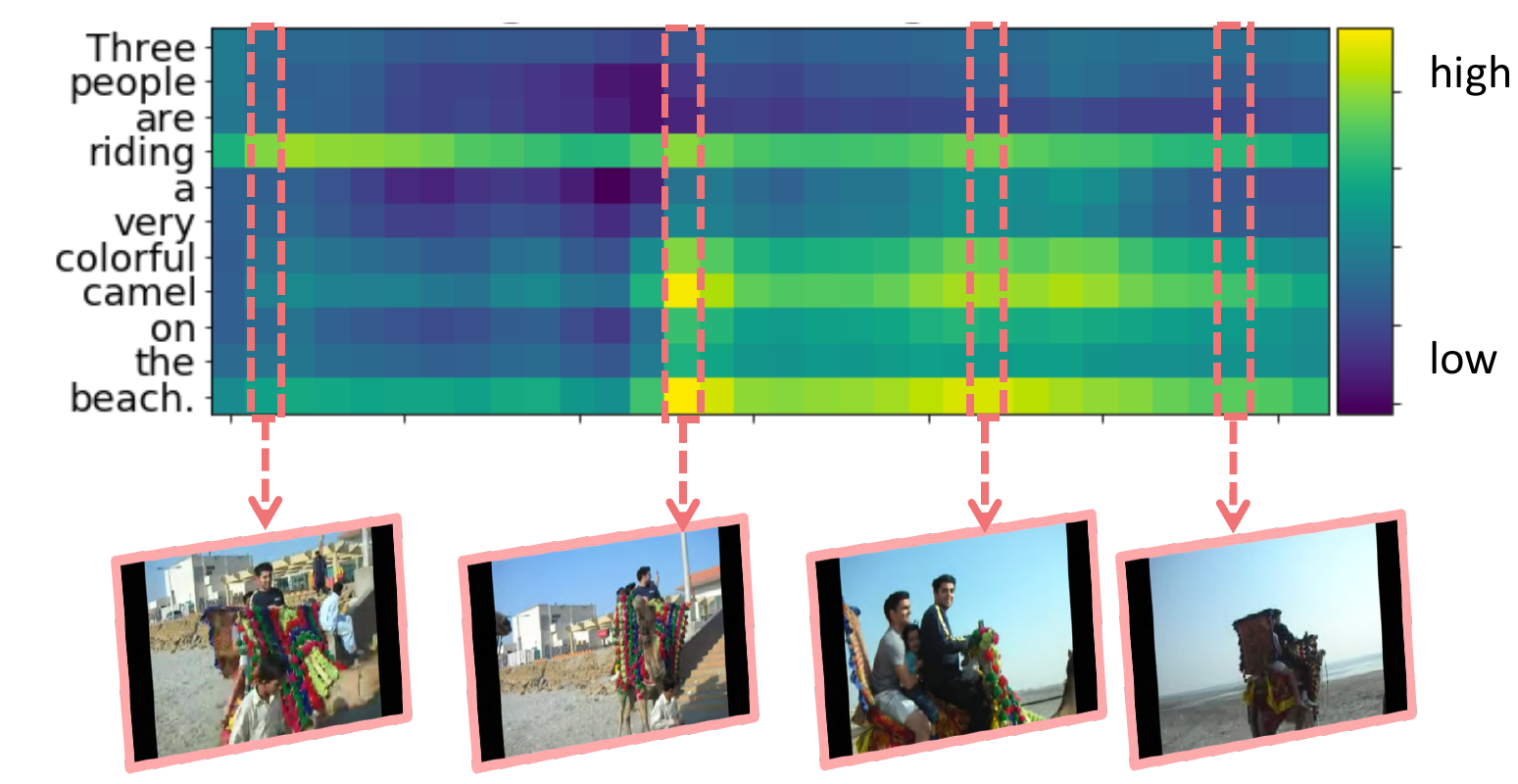} \label{fig:supp_att_vis_b}}

    \vspace{.1cm}
    \caption{\textbf{Visualization of graph matching attention.} We visualize the \textit{Matching Edge} of the graph matching layer. Correspondence between video snippets and query tokens can be evaluated through the heat-map.
    }
    \label{fig:supp_att_vis}
\end{center}
\end{figure}

In Fig.~\ref{fig:supp_att_vis}, we plot the \textit{Matching Edge} weights (before $\mathrm{SoftMax}$) for two video-query pairs, 
where the \textit{Matching Edge} weights are used to measure the similarity between video snippets and language tokens.
In graph convolutions, a \textit{Matching Edge} propagates more information if its weight is high, and vice versa. 

In Fig.~\ref{fig:supp_att_vis_a}, we show the grounding result for a 2 minutes accordian tutorial, with associated query: ``She is holding an accordian as she talks''. It can be observed from the blue-yellow heat-map that high scores are assigned to the words ``holding'', ``accordian'', and ``talks'', which are the most discriminative tokens for the query localization.
Below the heat-map, we visualize the snippets of the video. The unrelated snippets (first and last) are associated with low scores. Conversely, more relevant snippets (central ones) have higher \textit{Matching Edge} weights. This entails that the algorithm is successfully correlating important language cues with relevant video cues when performing the graph matching operation.  

Similarly, Fig.~\ref{fig:supp_att_vis_b} shows the result for a 22 second camel riding video, for which the associated query is: ``Three people are riding a very colorful camel on the beach.'' The heat-map highlights the keywords: ``riding'', ``colorful camel'', and ``beach'', which are relatively more informative in the query sentence. Interestingly, the word ``riding'' is always associated with high attention weights, and a visual inspection confirms that the action happens throughout the whole video. This showcases that our VLG-Net can successfully learn semantic video-language matching.
If we focus on the first two snippets of Fig.~\ref{fig:supp_att_vis_b}, we can see that both have associated high scores with the word ``riding''. However, given the smaller field of view of the first frame, only the second frame contains a more distinguishable camel. In fact, for this particular frame, we observe a high weight score for the words ``colorful'' and ``camel''. Moreover, the context of ``beach'' can be learned from all the last three snippets.


\begin{figure}[!ht]
\captionsetup[subfigure]{labelformat=empty}
\begin{center}

    \subfloat[(a) Learnable self-attention.]{ \includegraphics[trim={0cm 0cm 0cm 0cm},width=0.75\linewidth,clip]{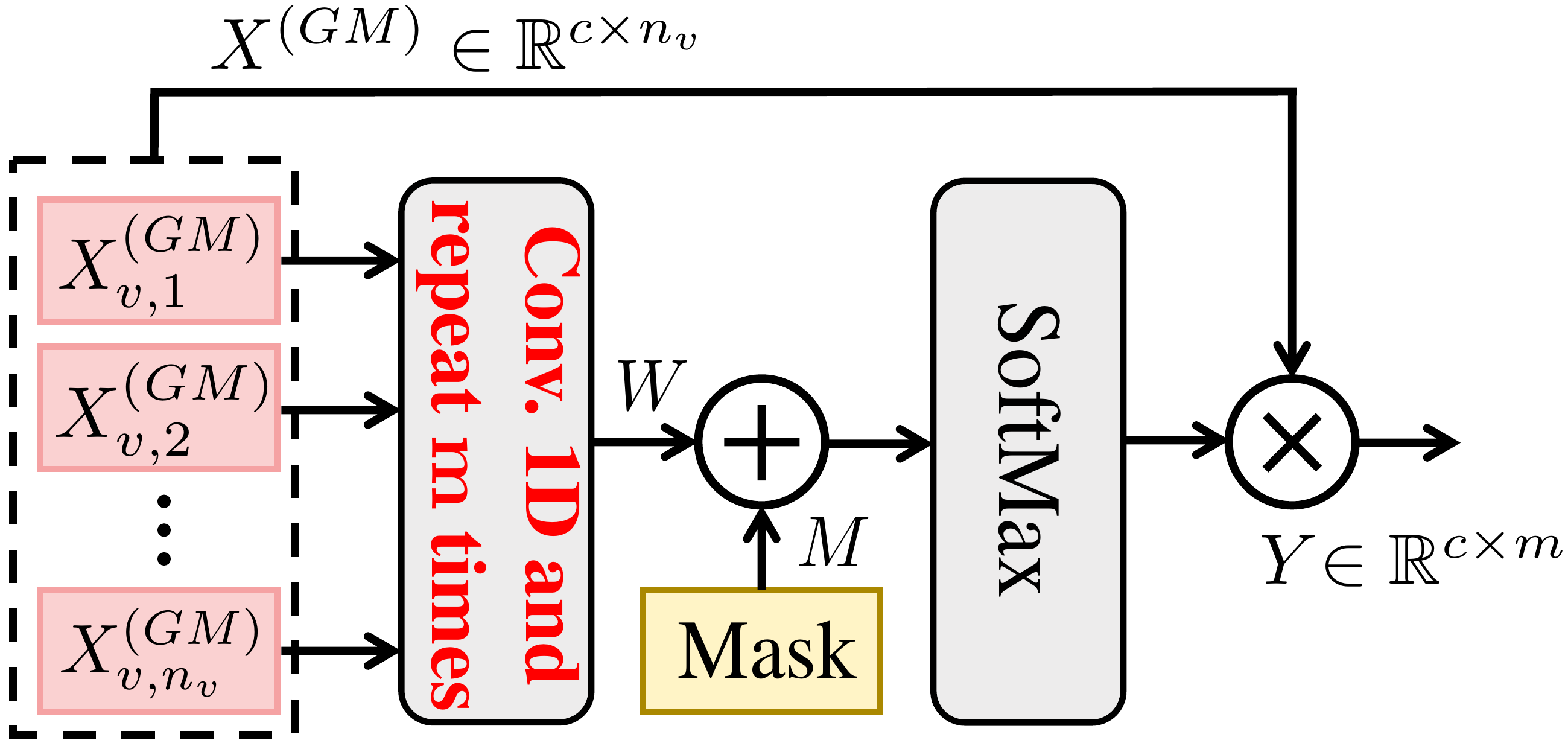} \label{fig:att_self}}\\
    
    \subfloat[(b) Cross-attention.]{\includegraphics[trim={0cm 0cm 0cm 0cm},width=0.87\linewidth,clip]{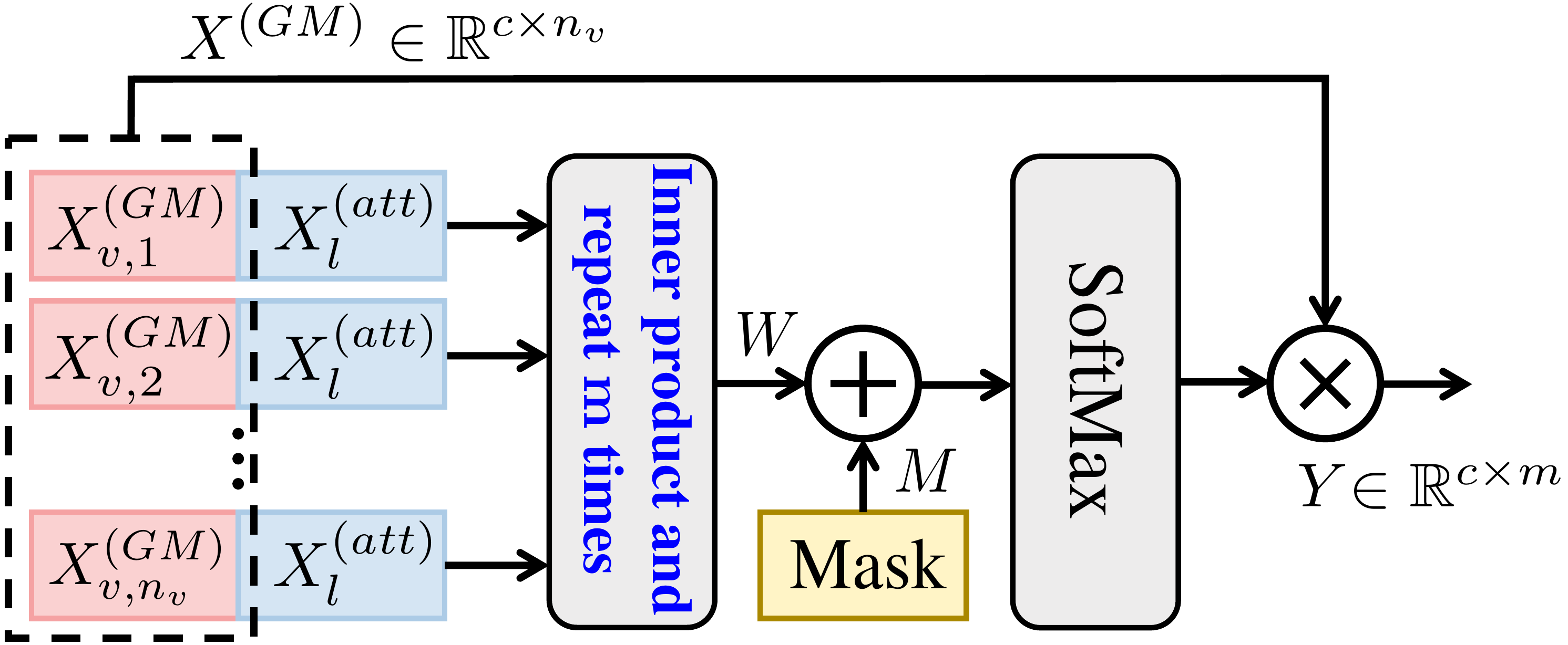} \label{fig:att_cross}}

    \subfloat[(c) Learnable cross-attention.]{\includegraphics[trim={0cm 0cm 0cm 0cm}, width=0.87\linewidth, clip ]{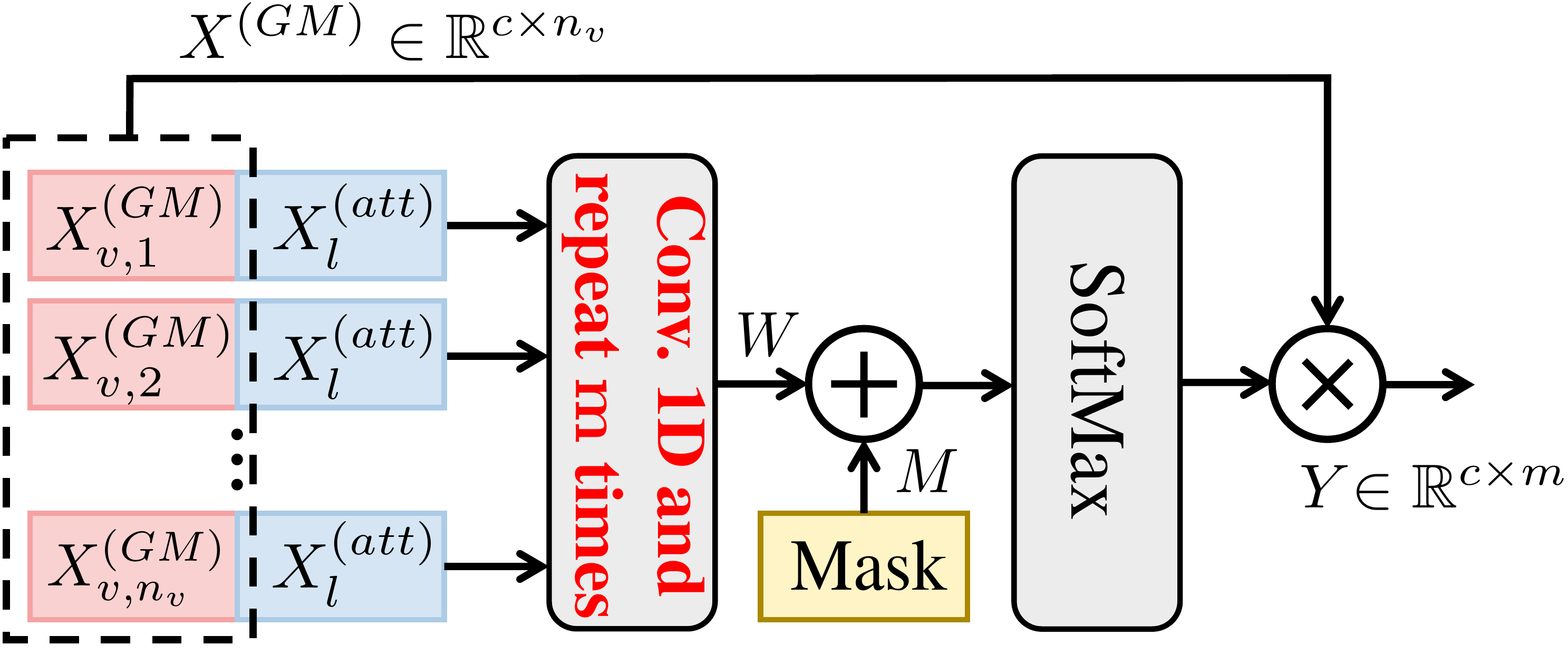} \label{fig:att_learnable_cross}}

    \vspace{.1cm}
     \caption{\textbf{Masked attention pooling.} Inputs are video nodes $X_v^{(GM)}$ from the graph matching layer and the query embedding $X_l^{(att)}$ computed through self-attention pooling atop the graph matching output. The output $Y$ represents all moment candidates.}
    \label{fig:att}
    
\end{center}
\end{figure}

\subsection*{Ablation of Masked Attention Pooling}
As presented in the main paper, three different implementations of attention for moment pooling operation have been tested. They differ for inputs and operations to achieve the attention scores. Learnable self-attention (Fig.~\ref{fig:att_self}), only relies on the fused features of video and language modalities, which are the output of the graph matching layer, while the cross-attention and learnable cross-attention configurations (Fig.~\ref{fig:att_cross} and~\ref{fig:att_learnable_cross}) also involve a global sentence representation $X_l^{(att)}$ in the process. (See Sec. 3.5 of the paper for more details.) 
We compare the performances of the three different implementations in Tab~\ref{tab:abl_att}.

Following the ablation settings in our main paper, we focus on R@1 IoU0.5 and R@5 IoU0.5 for TACoS dataset.
We find that the cross-attention setup leads to the lowest performance. Conversely the learnable cross-attention configuration instead, obtains the best performance. Therefore we adopt this configuration as default in the main paper. 

Interestingly we notice that the learnable self-attention setup can achieve relatively high performance. This can be motivated by the intuition that our graph matching layer can effectively fuse the video and language modalities, and by relying on those enriched features only, can we obtain a good representation of the moment's feature. However, involving a global language representation for guiding the moment creation from the enriched snippets features has been shown to yield the best results. 

\begin{table}[!t]
    \centering 
\setlength{\tabcolsep}{3pt}
\renewcommand{\arraystretch}{1.1} 
\resizebox{\linewidth}{!}{%
\scalebox{1.0}{
\begin{tabular}{l|ccc} 
\toprule
&  Learnable      & \multirow{2}{*}{Cross-attention}  & Learnable   \\
&  self-attention &                                   & cross-attention  \\
\midrule
R@1 IoU0.5 & $29.87$ & $16.62$  & $\mathbf{34.19}$    \\
R@5 IoU0.5 & $50.24$ & $40.14$  & $\mathbf{56.56}$    \\
\bottomrule                    
\end{tabular}
}
}
\vspace{.1cm}
\caption{\label{tab:abl_att}{\bf Ablation of masked attention pooling implementations.} The experimental results show that the cross-attention setup leads to sub-optimal performance. Instead, the learnable cross-attention configuration obtains the best performance.}
\end{table}

\begin{table}[!b]
    \renewcommand\thetable{8}
\centering 
\resizebox{\linewidth}{!}{%
\scalebox{1.0}{
\begin{tabular}{l|c|ccc|c} 
\toprule
Dataset             &   Num.   & \multicolumn{3}{c|}{Video-Sentence pairs} & Vocab.         \\ 
                    &   Videos & train & val & test                       & Size    \\ 
\midrule
ActivityNet Captions~\cite{Krishna_2017_ICCV} & 14926   & 37421  & 17505  & 17031 & 15406          \\
TACoS~\cite{TACoS_ACL_2013}                   & 127     & 10146  & 4589   & 4083   & 2255           \\
DiDeMo~\cite{Hendricks_2017_ICCV}             & 10642   & 33005  & 4180   & 4021   & 7523           \\
\midrule
Charades-STA~\cite{Gao_2017_ICCV}             & 6670    & 12404  & $-$      & 3720      & 1289           \\
\bottomrule
\end{tabular}
}
}
\vspace{.1cm}
\caption{{\label{tab:datasets2}{ \bf Datasets statistics}. Same as Table 1 in main paper, reported in Supplementary Material for completeness. }}

\end{table}

\subsection*{Charades-STA}

\begin{table}[!t]
    \centering 
\resizebox{0.8\linewidth}{!}{%
\scalebox{1.0}{
\begin{tabular}{l|cc} 
\toprule
                    & \multicolumn{2}{c}{Sentence's lengths}       \\
Dataset             &   Avg.  & Std.                               \\ 
\midrule
Activitynet-Captions~\cite{Krishna_2017_ICCV} & $14.4$   & $6.5$       \\ 
TACoS~\cite{TACoS_ACL_2013}                   & $9.4$    & $5.4$      \\
DiDeMo~\cite{Hendricks_2017_ICCV}             & $8.0$    & $3.4$      \\
\midrule
Charades-STA~\cite{Gao_2017_ICCV}             & $7.2$    & $1.9$       \\
\bottomrule
\end{tabular}
}
}
\vspace{.2cm}
\caption{\label{tab:datasets3}{ \bf Language annotations statistics}. We report average length (measured in number of tokens) and standard deviation for queries in each dataset. Statistics are computed considering every split for each dataset.  }

    \vspace{-0.4cm}
\end{table}

Based on the results obtained from Activitynet-Caption, TACoS, and DiDeMo, our method can theoretically achieve state-of-the-art performance in the Charades-STA dataset. However, we choose not to evaluate VLG-Net on this dataset because of the following observations.

(1) This dataset is characterized by the \textbf{smallest vocabulary size} and \textbf{shortest language annotation} with respect all others datasets (see Tab.~\ref{tab:datasets2} and Tab.~\ref{tab:datasets3})
For example, its vocabulary contains $43\%$ less unique words with respect to TACoS~\cite{TACoS_ACL_2013}, $83\%$ with respect to DiDeMo~\cite{Hendricks_2017_ICCV}, and $92\%$ with respect to Activity-Captions~\cite{Krishna_2017_ICCV}. This fact can potentially hamper the development of successful methods and reduce the applicability to a real-world scenario where \textbf{users might use a richer vocabulary} when querying for moments.
Given the great importance of the language for the task at hand, it's diversity in terms of unique tokens' number, and sentence lengths are important factors. This suggests that Charades-STA is less favourable for evaluating the video-language grounding task.

(2) Charades-STA has the \textbf{smallest number of video-query pairs} (16124) with respect to all other datasets (See Tab~\ref{tab:datasets2}).
As deep learning methods benefit from a large amount of annotated data, the reduced number of training/testing samples makes the dataset less suited for deep-learning approaches. 

(3) Most importantly, Charades-STA \textbf{lacks an official validation split.}
In machine learning applications, the validation set is mandatory for hyper-parameters search, while the test set is adopted for evaluating the generalization capabilities of a given method to previously unseen data. Given the absence of a validation set, nor a widely accepted procedure for selecting the best models during the development phase, some might use the test set for tuning the hyper-parameters, therefore, harming the measurement of generalization performance. 
The goal of research is to develop tailor-made solutions for specific problems rather than finding the hyper-parameters that can fit the test set best.
A conservative researcher could attempt at using the training set (or part of it) as a synthetic validation split. However, this could lead the model to overfit on the specific set of samples. Other methods could be potentially applied (\eg cross-validation), yet no previous work mentioned the adoption of such techniques. 

For all these reasons we can conclude that, despite the popularity of Charades-STA as benchmark for the language grounding in video task, we decide not to evaluate our method on it.

\end{document}